\documentclass[conference]{IEEEtran}
\usepackage{graphicx}

\makeatletter
\def\footnoterule{\relax%
  \kern-5pt
  \hbox to \columnwidth{\hfill\vrule width 0.8\columnwidth height 0.4pt\hfill}
  \kern4.6pt}
\makeatother

\usepackage{cite}
\usepackage{siunitx}
\usepackage{multirow}
\usepackage{xcolor}
\definecolor{redcolor}{rgb}{1, 0, 0}
\usepackage{subfiles}
\usepackage{makecell}
\usepackage{color,soul}
\usepackage{amsmath}
\usepackage{multirow}
\usepackage{mathtools}
\usepackage{csquotes}
\usepackage{esvect}
\usepackage{stfloats}   %for two column wide figure and table
\usepackage{float}

\usepackage{multirow}

\IEEEoverridecommandlockouts
\begin{document}

% set the distance before and after figure/eqution
\abovedisplayskip=0pt
\abovedisplayshortskip=0pt
\belowdisplayskip=0pt
\belowdisplayshortskip=0pt

%\abovecaptionskip=0pt
%\belowcaptionskip=-10pt
\setlength{\belowcaptionskip}{-10pt}

\title{Experimental Demonstration of Neuromorphic Network with STT MTJ Synapses
}
\author{\IEEEauthorblockN{
Peng Zhou\IEEEauthorrefmark{1},
Alexander J. Edwards\IEEEauthorrefmark{1},
Fred B. Mancoff\IEEEauthorrefmark{2},
Dimitri Houssameddine\IEEEauthorrefmark{2},
Sanjeev Aggarwal\IEEEauthorrefmark{2},\\
Joseph S. Friedman\IEEEauthorrefmark{1}
}
\IEEEauthorblockA{\IEEEauthorrefmark{1}The University of Texas at Dallas, Richardson, Texas, USA. Email: joseph.friedman@utdallas.edu}
\IEEEauthorblockA{\IEEEauthorrefmark{2}Everspin Technologies, Inc., Chandler, Arizona, USA.}
%\thanks{DISTRIBUTION STATEMENT C.  Distribution authorized to U.S. Government agencies and their contractors.}
}

\maketitle
% make the title


\begin{abstract}
We present the first experimental demonstration of a neuromorphic network with magnetic tunnel junction (MTJ) synapses, which performs image recognition via vector-matrix multiplication.  We also simulate a large MTJ network performing MNIST handwritten digit recognition, demonstrating that MTJ crossbars can match memristor accuracy while providing increased precision, stability, and endurance.
%if Keywords are used, end abstract text with \\
\\
\end{abstract}
\renewcommand\IEEEkeywordsname{Keywords}

% \fi
%\IEEEpeerreviewmaketitle

\section{Introduction}

One of the primary hardware costs incurred by artificially-intelligent neural networks is due to vector-matrix multiplication (VMM), which is generally performed with floating-point binary numbers and computed exactly with cumbersome multiplication computations which scale with the number of entries in the matrix; \textit{i.e.}, $O(N^2)$. As neural networks are robust to imprecision, there is therefore an opportunity to improve neural network efficiency through the approximate analog calculation of VMM.

\begin{figure}[!t]
	\centering
	\includegraphics[width=0.75\columnwidth]{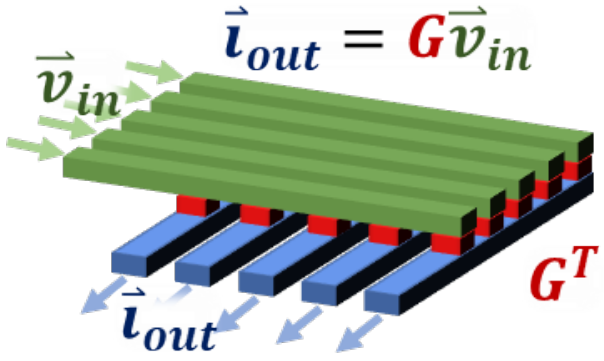}
	\caption{VMM with a memristor crossbar array.  The matrix values are encoded in the memristor conductances, \textbf{\textit{G}}.  The input vector is converted to an array of voltages fed into the row lines, $\vec{\textbf{\textit{v}}}_{in}$. When the column lines are grounded, the resulting current vector through them is the output of the VMM operation, $\vec{\textbf{\textit{i}}}_{out}$. This current can be sensed, and an analog to digital converter used to convert to digital signals.}
	\label{fig:fig1}
\end{figure}

\begin{figure}[]
	\centering
	\includegraphics[width=0.6\columnwidth]{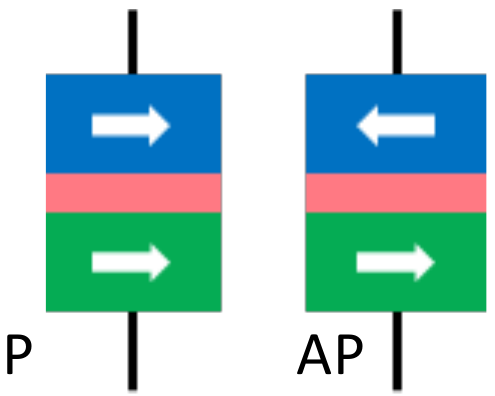}
	\caption{STT-MTJs.  The resistance of the two-terminal STT-MTJ depends on the relative magnetizations of the free (blue) and fixed (green) layers, having a low (high) resistance when they are in (anti-) parallel orientations.  The fixed layer magnetization never changes, whereas the free layer can be switched between the two stable states.  In an STT device this is accomplished with a short, high-amplitude burst of current.}
	\label{fig:fig2}
\end{figure}

Neuromorphic computing aims to mimic both the functionality and structure of the human brain through the interconnection of biomimetic neurons and synapses. Non-volatile analog resistive memory devices appear to naturally mimic the behavior of neurobiological synapses, and can be used to perform VMM by taking advantage of Ohm's and Kirchhoff's laws to efficiently perform these calculations in terms of voltage and current (Fig. \ref{fig:fig1}). Most research in this area has focused on crossbars composed of memristors and phase change memory (PCM) \cite{Cai2019}, in which the device resistance can be written and erased via voltage pulses across the device. However, these devices suffer from several major challenges which limit their utility for VMM crossbars, including 
\begin{itemize}
    \item stochastic writing that impedes precise setting to a desired resistance state,
    \item weights drifting over time through stochastic processes,
    \item limited endurance, and
    \item compatibility challenges with modern CMOS processes.
\end{itemize}
\noindent The magnetic tunnel junctions (MTJs) shown in Fig. \ref{fig:fig2} provide solutions to the limitations of memristors and PCM, but only exhibit binary resistance states. Importantly, it was recently proposed that stochastic MTJ switching enables analog neuromorphic behavior with MTJs \cite{French-Stochasitc-MTJ-Network}, though this has not yet been experimentally demonstrated.

Here we present the first experimental demonstration of a neuromorphic network using MTJ synapses for VMM.  While spintronic devices have been demonstrated for other neuromorphic functions -- \textit{e.g.} spin-torque nano oscillators \cite{Torrejon2017-Neuromorphic-STNOs}, domain wall MTJs \cite{Siddiqui2020-DWMTJ}, and stochastic p-bits \cite{FriedmanMTJ-P-BIT} -- they have not previously been demonstrated as adjustable synapses in a neuromorphic network.  This paper reports the experimental demonstration of a small binarized neuromorphic network with MTJ synapses, greatly advancing the development of highly-efficient, robust hardware amenable to neural applications.

\section{Background: Magnetic Tunnel Junction Synapse for Neuromorphic Computing}

The switching between the two spin-transfer torque (STT)-MTJ magnetic states is intrinsically stochastic, with a switching probability dependent on the write pulse voltage and duration time.  Vincent \textit{et al.} \cite{French-Stochasitc-MTJ-Network} proposed that the stochastic switching of STT-MTJs can emulate analog synapse behavior in a neuromorphic system, which can overcome the limitations of binary weights provided by MTJs. Further, their simulation results indicate strong robustness to device mismatch and variations, thus providing a pathway for neuromorphic computing with binary MTJs.

\section{Vector-Matrix Multiplication with Neuromorphic MTJ Synaptic Networks}

\begin{figure}[]
	\centering
	\includegraphics[width=0.65\columnwidth]{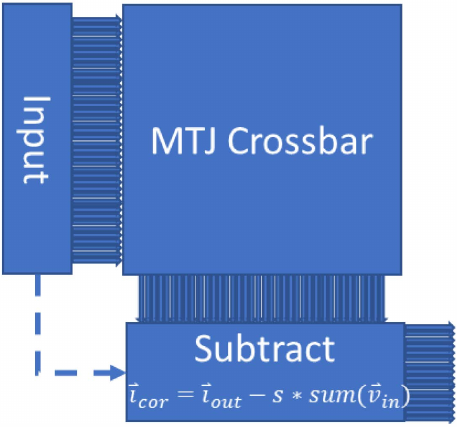}
	\caption{Binary VMM with an MTJ crossbar array.  As the MTJs have binary positive conductance states, a correction step is used at the crossbar output.  This correction step subtracts a scaled version of the input vector according to equation (2).  This step enables a wide range of weights including zero-valued or negative weights.}
	\label{fig:fig3}
\end{figure}

One of the primary challenges impeding the use of MTJs in a VMM crossbar array is the low on-off resistance ratio, limiting the ability to implement synaptic weights of zero.  As shown in Fig. \ref{fig:fig3}, we propose that this challenge can be overcome by including simple logic that efficiently resolves this issue, enabling zero-valued and negative weights with minimal additional logic.

\begin{figure}[]
	\centering
	\includegraphics[width=1\columnwidth]{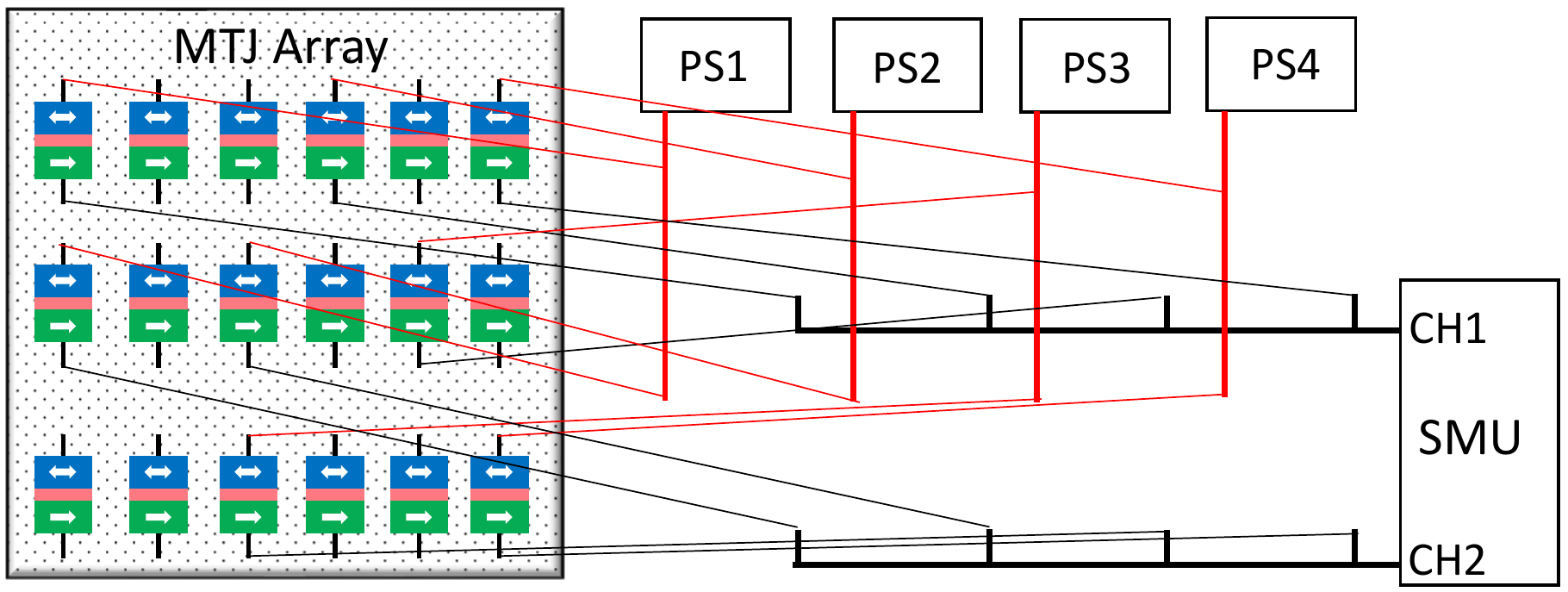}
	\caption{Schematic of the experimental setup. The input voltages are provided by the power supplies (PS1-4), and the output currents are read by the two channels of the source meter. For each MTJ, the blue layer represents the free layer, and the green layer represents the fixed layer.  Note that the MTJ layout here is an example and does not match our experimental chip.}
	\label{fig:fig4}
\end{figure}

\begin{figure}[]
	\centering
	\includegraphics[width=1\columnwidth]{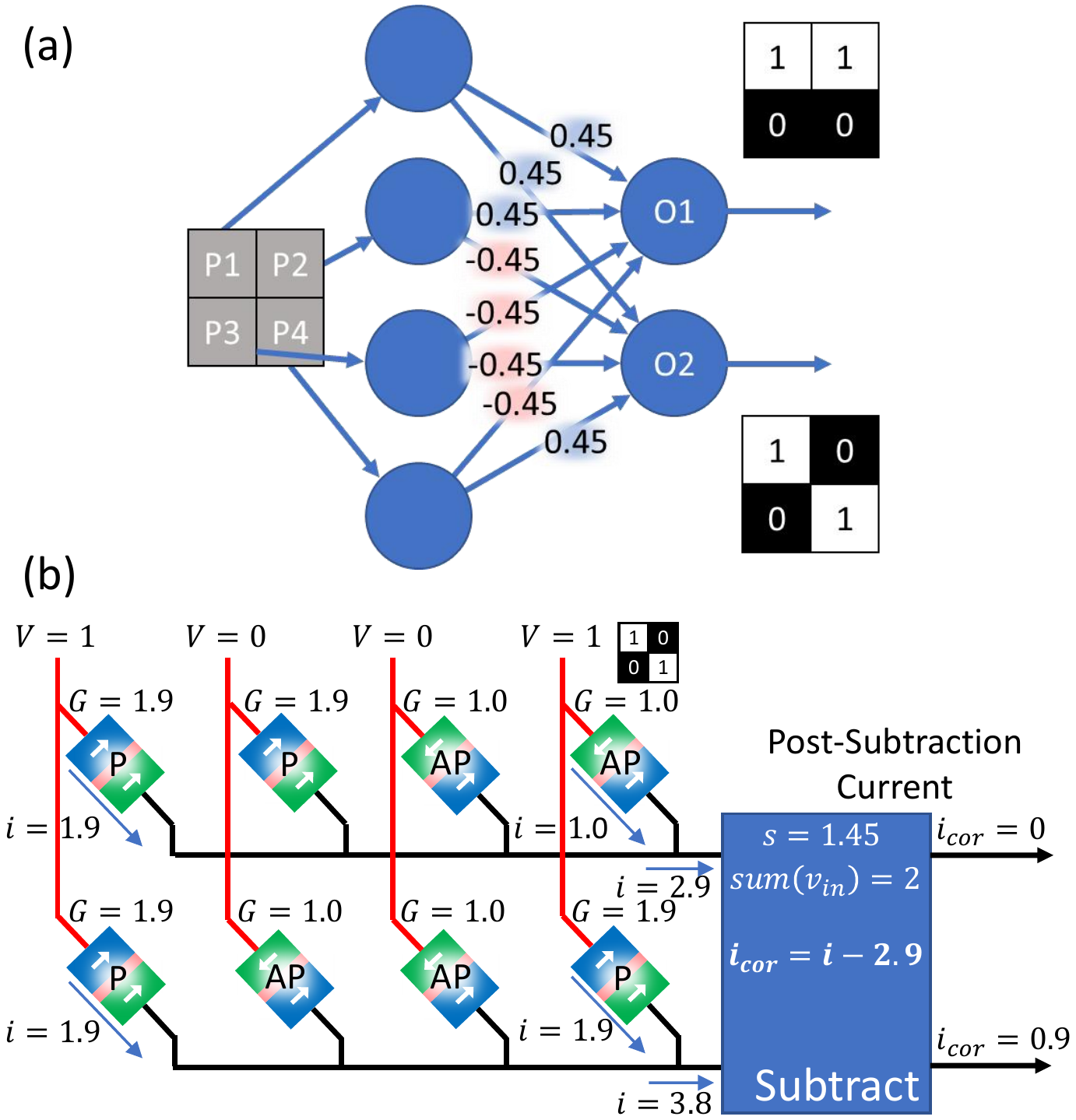}
	\caption{(a) Image recognition using a small neural network.  A 4x2 single layer neural network capable of classifying 2x2 images.  For classification, the input image is vectorized and fed into the network which compares it pixel-by-pixel to the output classes.  A weight of +0.45 (-0.45) is used for expected white (black) pixels. (b) Trained MTJ neural network.  Physical quantities are normalized such that  $v_r = 1$ and $G_{AP} = 1$ .  Given this configuration, the output ranges between $I = \pm0.9$, with positive values corresponding to a greater match between the input image and the target.  In this case, the image (1, 0, 0, 1) is fed into the network matching target image 2 identically and target image 1 neutrally.}
	\label{fig:fig5}
\end{figure}

Let the parallel (P) and anti-parallel (AP) MTJ conductances be denoted by $G_P$ and $G_{AP}$, respectively, within an MTJ crossbar array $\mathbf{G}$.  Let $v_{in}$ be a binarized input voltage vector with values $0$ and a fixed `1' value of $v_{r}$.  The column output current $\mathbf{i_{out}}$ can then be computed as 

\begin{equation}
    \mathbf{i_{out}} = \mathbf{G} \cdot \mathbf{v_{in}} = \sum_{n \mid \mathbf{v_{in}}_{,n} \ne 0} v_{r} * \mathbf{G_n}
\end{equation}

\noindent where $\mathbf{G_n}$ is the $n^{th}$ column of the conductance matrix. We propose implementing negative weights by subtracting the effect of the AP conductances.  Let $s = (G_P + G_{AP}) / 2$ be the weight adjustment factor and correct $\mathbf{i_{out}}$ with 

\begin{equation}
\begin{multlined}
    \mathbf{i_{cor}} = \mathbf{i_{out}} - s*sum(\mathbf{v_{in}}) = \\ = \mathbf{i_{out}} - s * nnz(\mathbf{v_{in}}) * v_{r}
    = \sum_{n\mid \mathbf{v_{in}}_{,n} \ne 0} v_{r} * (\mathbf{G_n} - s),
    \end{multlined}
\end{equation}

\noindent where $nnz(A)$ is the number of non-zero elements of $A$. This effectively shifts each weight to be either $-(G_P-G_{AP})/2$ or $+(G_P-G_{AP})/2$.  $\mathbf{i_{cor}}$ may be computed with very simple logic at the output of the VMM, and different values of $s$ may be chosen to enable different weights (including zero-valued weights). Thus, with this correction step, the weights are not constrained to the P and AP resistances and may be chosen arbitrarily, opening wide the application space for MTJ synaptic crossbars. The following demonstration uses this correction step to emulate weights symmetric about 0.

\begin{figure}[!t]
	\centering
	\includegraphics[width=1\columnwidth]{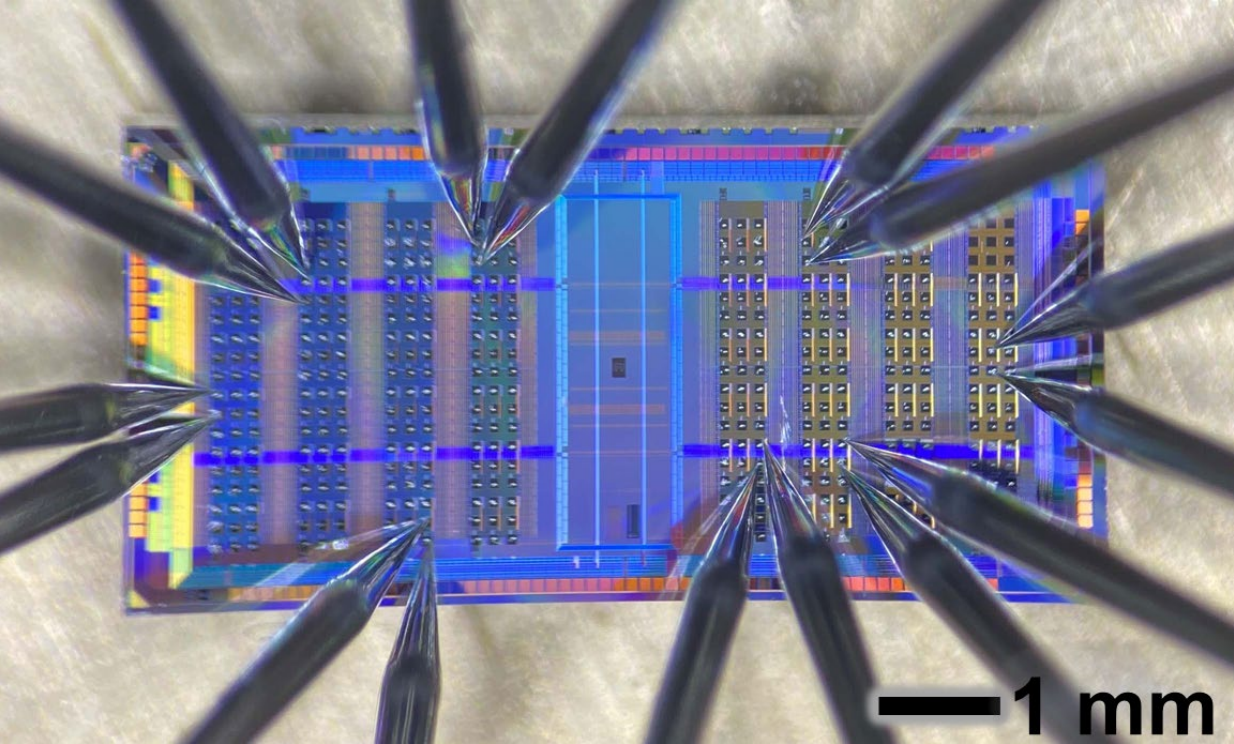}
	\caption{Experimental setup.  Eight MTJs from the depicted MTJ array were probed and connected as in Fig 4.}
	\label{fig:fig6a}
\end{figure}

\begin{figure*}[!b]
    \centering
    \includegraphics[width=0.7\textwidth]{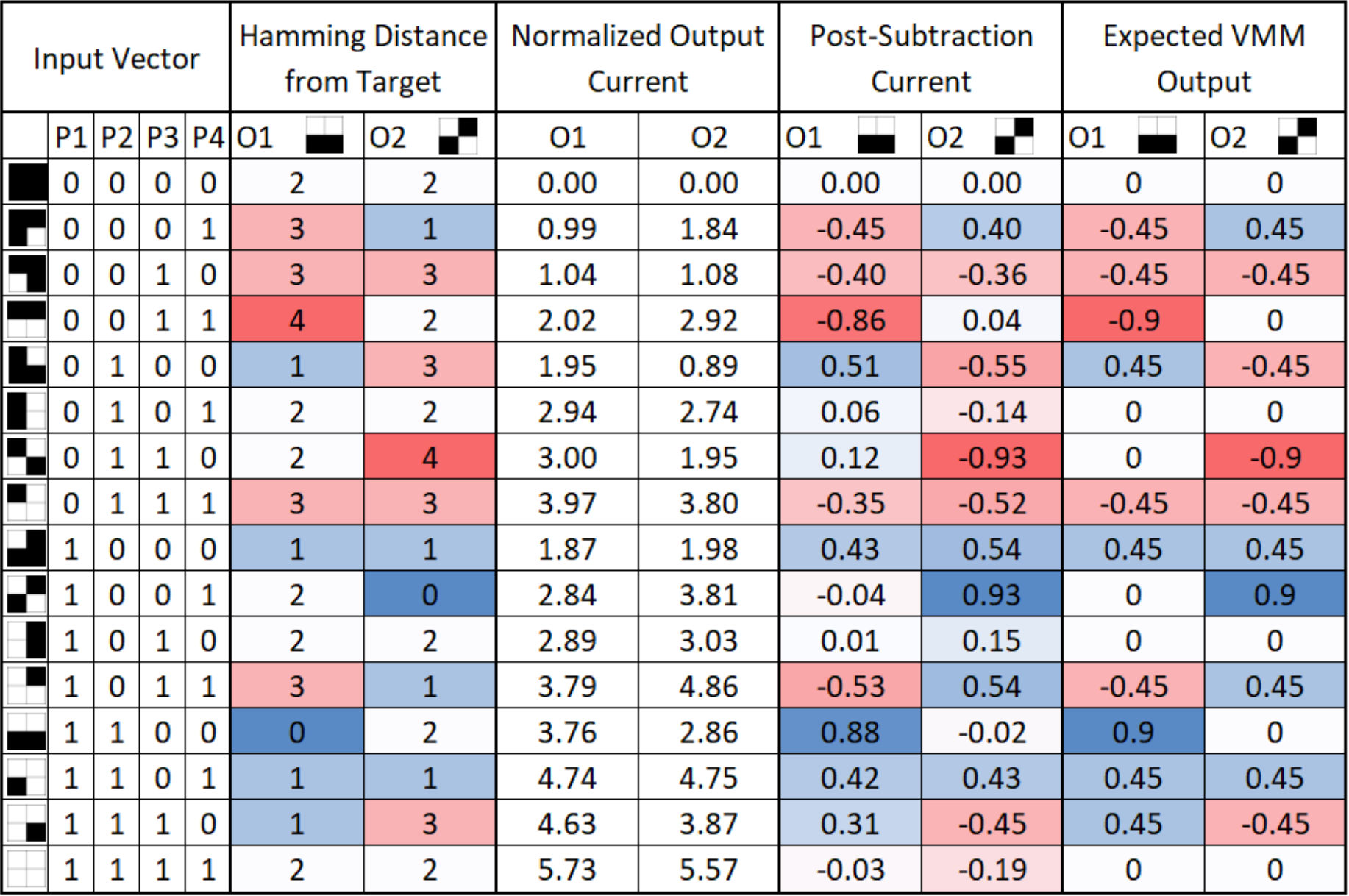}
    \caption{Experimental Results.  All possible input images were presented to the MTJ VMM network.  The Hamming distance between the input and target images was calculated.  A color scale from blue to red is used to show image alignment with blue (red) indicating maximum (minimum) alignment.  DC input voltages were fed into the network and the output currents were measured and normalized by the average current through the AP MTJs.  The subtraction factor described in Sec III was calculated to be 1.44 after normalization and was subtracted from the measured output currents giving the VMM results shown.  The same color scheme is used for this output demonstrating successful VMM experimental calculation for all input combinations.\vspace{-10px}}
    \label{fig:fig6b}
\end{figure*}

\section{Experimental Setup}

Our experiment demonstrates a 4x2 neuromorphic VMM accelerator, with eight two-terminal MTJ synapses as shown in Fig. \ref{fig:fig4}. The four input voltages are provided by four power supplies (PS) and the two output currents are read by the source measuring unit (SMU). Each MTJ needs two probes to connect the top and bottom electrodes individually, thereby requiring a total of 16 probes for this 4x2 neuromorphic network.

%We implemented the network above using an MTJ array with the ... probe station and 16 probes.  MTJs were written and sensed in the network.  Fig. 5 (Schem)  Fig. 6 (Picture)

%%%The MTJs are rotated to show the wire connections that prevent the stochastic switching during the test. %%%

\section{Task \& Results}

To evaluate the functionality of the proposed system, we tested the system on the two pixel by two pixel supervised image classification task shown in Fig. \ref{fig:fig5}(a).  In this task, four input pixels are fed into the network, which is tasked with recognizing two target images.  The network must calculate the Hamming distance between the input and target images, to which a threshold may be applied to identify images that are identical or sufficiently similar to the target image. The synapses in the network were trained offline to weights of either +0.45 or -0.45, corresponding respectively to the P and AP MTJ resistance states.

The resistance of each MTJ is initialized to either the P or AP state based on the network training. This is shown in Fig. \ref{fig:fig5}(b), which depicts the electrical response for one particular input image, as well as the crossbar output both before and after the subtraction logic.  
%All MTJs are connected in a manner that prevents STT switching during testing, though this precaution would be unnecessary in an integrated system with read voltages controllably applied for very brief periods.  
Fig. \ref{fig:fig6a} depicts our 16 probes contacting the eight MTJs selected for the experimen.

All 16 input voltage combinations are fed into the MTJ array and the output current is measured; currents normalized by the average AP state current are presented in Fig. \ref{fig:fig6b}. The output current variations result primarily due to power supply voltage variation and differences in probe connectivity. Significantly decreased variation is expected for future neuromorphic networks with MTJ arrays directly integrated with the peripheral CMOS circuits.

Fig. \ref{fig:fig6b} includes the post-subtraction results of the normalized current using (2). By setting proper thresholds, the post-subtraction results can be categorized into five distinct bins. These post-subtraction currents show high fidelity to the expected VMM results for this four-pixel input task, demonstrating robustness to device and testing imprecision.  This circuit thus successfully performs the VMM.

\begin{figure}[!t]
	\centering
	\includegraphics[width=0.85\columnwidth]{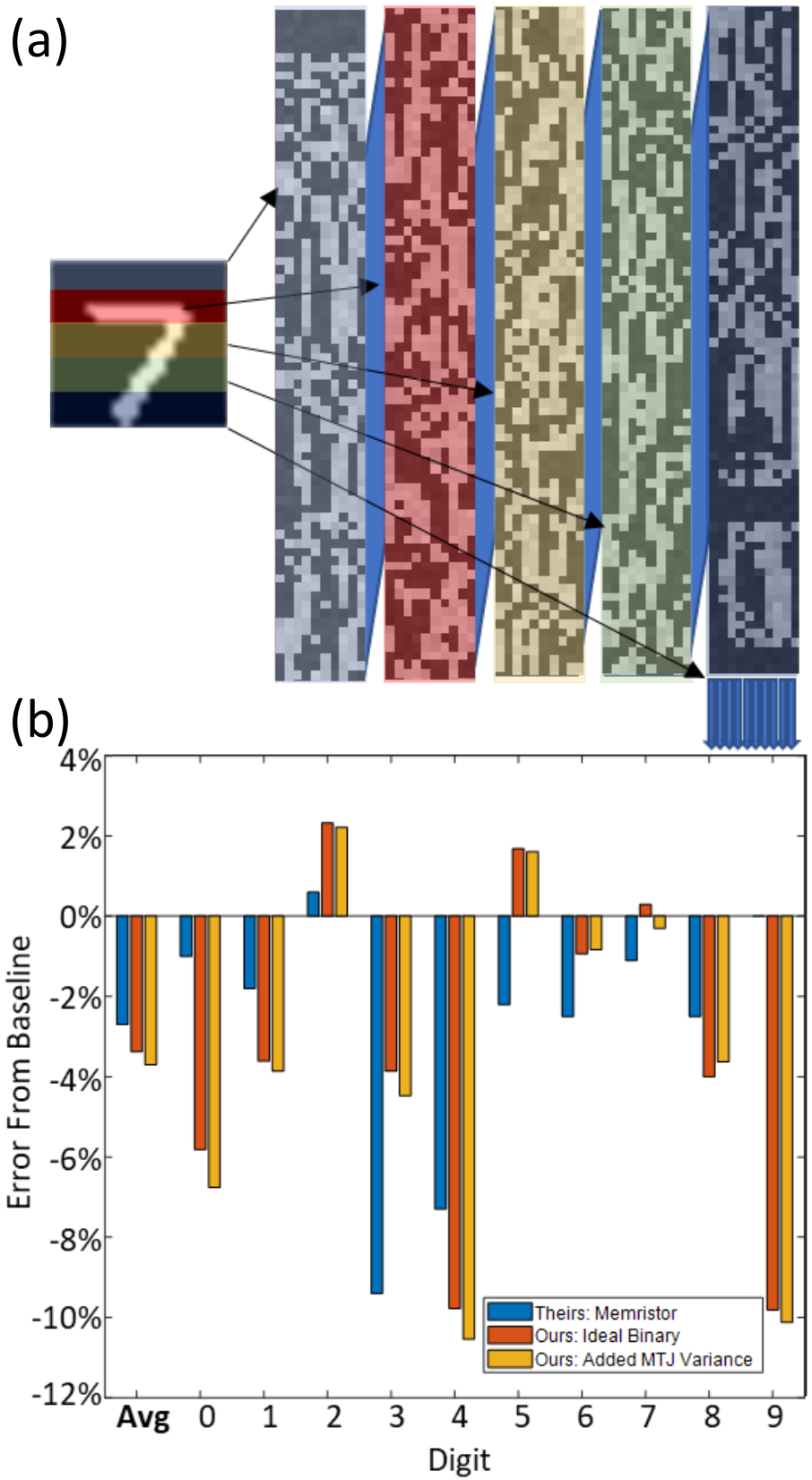}
	\caption{(a) Trained binary MTJ weights for single-layer 380x10 MNIST handwritten digit recognition network with softmax activation at the output layer.  Light (dark) colors correspond to (anti)-parallel resistances. (b) Simulated MNIST handwritten digit recognition neural network with MTJs compared with memristor network [6].  The 3.7\% accuracy loss between our baseline (analog) network and the MTJ network is comparable to the the 2.7\% loss between their baseline and the memristor network.}
	\label{fig:fig7}
\end{figure}

\section{MTJ Crossbar Scaling}

To demonstrate this robustness, we trained and simulated a single-layer neuromoprhic network with MTJ synapses to perform the MNIST handwritten digit recognition task similar to the experiment performed in \cite{Memristor_MNIST}.  As in \cite{Memristor_MNIST}, images were scaled to be 19x20 pixels.  Our 380x10 single-layer network applies a softmax activation function output layer and was trained using the matrix pseudo-inverse.  Weights were then binarized according to a fixed threshold to map into an MTJ array.  No further weight optimizations were performed on the network.  A Monte Carlo simulation was performed in which device-to-device variations and tunneling magnetoresistance were chosen to match experimental values. The resultant weight matrix is depicted in Fig. \ref{fig:fig7}(a).

Fig. \ref{fig:fig7}(b) illustrates a comparison between the benchmark memristor system in \cite{Memristor_MNIST} and our simulated MTJ network.  Bars are drawn relative to the ideal analog neural networks in both works to depict loss in accuracy when imprecision is incorporated.  Binarization of the weights accounts for most of the accuracy loss, with MTJ variance adding 0.3\% loss; we expect that with binary weight optimization, the MTJ system will be more accurate than the memristor system. This is consistent with the large robustness to variations demonstrated in the simulations of  \cite{French-Stochasitc-MTJ-Network}.   MTJ device-to-device variation thus does not impede the scaling of neuromorphic networks to large sizes.

\section{Conclusions}

We report the first experimental demonstration of a neuromorphic network with MTJ synapses, and demonstrate that variance in MTJ resistance results in minimal accuracy loss when scaled to larger networks.   This work thus provides initial experimental proof that MTJ synaptic crossbars can perform VMM in a manner similar to memristor crossbars, while providing a pathway for increased precision, stability, endurance, and technology integration.

\section*{Acknowledgment}
This work is supported by Semiconductor Research Corporation (SRC) Task No. 2810.030 through UT Dallas’ Texas Analog Center of Excellence (TxACE).

\bibliographystyle{IEEEtran}
\bibliography{IEEEabrv,GOMACTech_conf}

\end{document}